\documentclass{article} 
\usepackage{nips13submit_e,times}
\usepackage{hyperref}
\usepackage{url}
\usepackage{cite}
\usepackage{graphicx}
\usepackage{float}
\usepackage{subfig}

\title{GPU Asynchronous Stochastic Gradient Descent to Speed Up Neural Network Training}

\author{
Tom Paine\\
University of Illinois at Urbana-Champaign\\
Urbana, IL \\
\texttt{paine1@illinois.edu} \\
\And
Hailin Jin \\
Adobe Research \\
San Jose, CA \\
\AND
Jianchao Yang \\
Adobe Research \\
San Jose, CA \\
\And
Zhe Lin \\
Adobe Research \\
San Jose, CA \\
\And
Thomas Huang \\
University of Illinois at Urbana-Champaign \\
Urbana, IL \\
}

\nipsfinalcopy 

\begin{document}

\maketitle

\begin{abstract}
The ability to train large-scale neural networks has resulted in state-of-the-art performance in many areas of computer vision. These results have largely come from computational break throughs of two forms: \textit{model parallelism}, e.g. GPU accelerated training, which has seen quick adoption in computer vision circles, and \textit{data parallelism}, e.g. A-SGD, whose large scale has been used mostly in industry. We report early experiments with a system that makes use of both model parallelism and data parallelism, we call \textbf{GPU A-SGD}. We show using GPU A-SGD it is possible to speed up training of large convolutional neural networks useful for computer vision. We believe GPU A-SGD will make it possible to train larger networks on larger training sets in a reasonable amount of time.
\end{abstract}

\section{Introduction}
Recently, large convolutional neural networks have achieved state-of-the-art results across many areas of computer vision including: character recognition \cite{ciresan2012multi}, object recognition \cite{le2011building, krizhevsky2012imagenet,2013arXiv1311.2901Z,donahue2013decaf}, and object detection \cite{girshick2013rich}. This is partly the result of larger datasets, e.g. the Imagenet Large Scale Visual Recognition Challenge (ILSVRC) \cite{deng2009imagenet} and accelerated training algorithms that can make use of the data. These approaches may be accelerated by using many CPUs \cite{zinkevich2010parallelized, agarwal2012distributed, niu2011hogwild, NIPS2012_0598}, or GPUs \cite{raina2009large,bergstra2010theano,collobert2011torch7,krizhevsky2012imagenet,ciresan2012multi}, and even many GPUs \cite{coates2013deep
}. We believe accelerating training further will result in more break throughs in computer vision.

We present experiments using a new system for accelerating neural network training, using \textbf{asynchronous stochastic gradient descent (A-SGD)} with many GPUs, which we call GPU A-SGD. We show that this system can be used to speed up training by several times, and explore how to best use GPU A-SGD to further speed up training. To benchmark our speed up, we use the pipeline found in\cite{krizhevsky2012imagenet}. We train a convolutional neural network on the ILSVRC 2012 dataset, which has 1000 classes, and 1.2 million images. Like that work, our network uses dropout \cite{hinton2012improving}, relu neurons \cite{nair2010rectified}, and is trained use data augmentation.

We will first review neural network training algorithms. And then highlight how our training algorithm differs from existing methods.

\section{Training neural networks}
A neural network can be seen as a large parameterized function. The parameters in this function can be learned through gradient descent style algorithms. In traditional gradient descent, the gradient of the objective function needs to be calculated over the entire dataset. The parameters are then updated with this gradient. This is repeated until convergence. There are two main issues with this approach: The dataset may be too large to fit into memory, and the gradient may take too long to compute.

When the dataset is too large, \textbf{stochastic gradient descent (SGD)} may be used. Here the gradient of the objective function is calculated over a small random partition of the dataset called a \textbf{minibatch}. The parameters are updated with this minibatch gradient, and a new minibatch is chosen. This process is repeated until convergence. This algorithm can be accelerated in two ways: speeding up the calculation of the minibatch gradient (\textbf{model parallelism}), and parallelization of the stochastic gradient descent steps (\textbf{data parallelism}).

\subsection{Model parallelism}
In many approaches, the structure of neural network computations is exploited to speed up the calculation of the minibatch gradient. This can be called model parallelism. This can be achieved using GPUs \cite{raina2009large,bergstra2010theano,collobert2011torch7,krizhevsky2012imagenet,ciresan2012multi}, distributed CPU approaches \cite{NIPS2012_0598}, or distributed GPU approaches \cite{coates2013deep
}. The distributed approaches have the added benefit that they can train models that are too big to fit in memory on a single device. In many cases, these models ignore parallelization of SGD, with \cite{NIPS2012_0598} being the notable exception. It's \textbf{DistBelief} technique makes use of both model parallelism, and data parallelism, which we will talk about more below.

One work \cite{coates2013deep} is similar to ours in that they experiment with many GPUs in a distributed framework to accelerate computation of very large models. Their work differs from ours because they \textbf{primarily focus on model parallelism} to train models too big to fit on a single device, especially for \textbf{unsupervised pre-training} of \textbf{locally-connected neural networks}. They are able to train the billion parameter model of \cite{le2011building}, using a significantly smaller number of nodes by leveraging consumer off-the-shelf GPUs and high-speed interconnect. While this line of research is very promising, these locally-connected, unsupervised models are not currently the top performing models on common computer vision benchmarks like ILSVRC. We believe our approach is complementary to theirs.

\subsection{Data parallelism}
Another method for speeding up training of neural networks is using distributed versions of stochastic gradient decent \cite{zinkevich2010parallelized, agarwal2012distributed, niu2011hogwild, NIPS2012_0598}. These methods can be called data parallel because they speed up the rate as which the entire dataset contributes to the optimization.

The data parallel part of the DistBelief \cite{NIPS2012_0598} model, (A-SGD) is especially interesting, because it is essentially many neural network models training independently, and occasionally communicating with a central parameter server to synchronize the overall effect for many distributed gradient updates. This makes it straight-forward to apply with various model parallel approaches. This model has also proved useful for computer vision problems, achieving state-of-the-art performance on a computer vision benchmark with 14 million images\footnote{Imagenet Fall 2011 release, not to be confused with the ILSVRC 2012, which is a subset of the Fall release.} \cite{le2011building}. While these methods may outperform single GPU based methods, by leveraging many more parameters, they operate at a very large scale (thousands of CPU cores).

\subsection{Our contribution: GPU A-SGD}

Our work also exploits both model parallelism, and data parallelism. We use GPUs for model parallelism, and A-SGD for data parallelism.
A-SGD is a subset of the DistBelief system described in \cite{NIPS2012_0598}. Our technique ignores their distributed CPU approach for model parallelism, and instead used GPUs to accelerate gradient computation. Multiple replicas of a model are used to optimize a single objective. Each model replica is trained using a GPU. This is achieved by extending the publicly available \textit{cuda-convnet} code\footnote{The original \textit{cuda-convnet} code is available at: https://code.google.com/p/cuda-convnet/} used in \cite{krizhevsky2012imagenet} to allow several GPU clients to communicate with a server. We use MPI for communication.

Each model requests updated parameters every $n_{fetch}$ steps, and sends updated gradient values every $n_{push}$ steps. In the DistBelief paper \cite{NIPS2012_0598} $n_{fetch} = n_{push} = 1$. This regime would not work well for GPUs, where the gradients are not usually communicated to the CPU after every minibatch. Typically the parameters are updated on the GPU for $n_{sync}$ steps before being copied to the CPU, where $n_{sync}$ can be large, e.g. 600. This is because there is additional overhead cost for transferring the parameters from the GPU to the CPU. This overhead can reduce the benefit for GPU accelerate gradient calculations. In our experiments we set $n_{fetch}=n_{push}=n_{sync}$. We experiment with different values of $n_{sync}$.

\section{Experiments}

To test GPU A-SGD, we train a convolutional neural network with the same architecture as described in \cite{krizhevsky2012imagenet} on the ILSVRC 2012 dataset. On a single NVIDIA Tesla K20X GPU this takes about 10.7 days.

We performed all experiments on the Blue Water supercomputer. It has over 4000 Nvidia Tesla K20X nodes, and has a Gemini high-speed interconnect. While we make use of a very high-performance machine, \cite{coates2013deep} notes that GPUs and high speed interconnect are now available off-the-shelf. All of our experiments are performed with 32 or less GPU nodes.

\begin{figure}[h]
    \centering
    \includegraphics[width=0.65\textwidth]{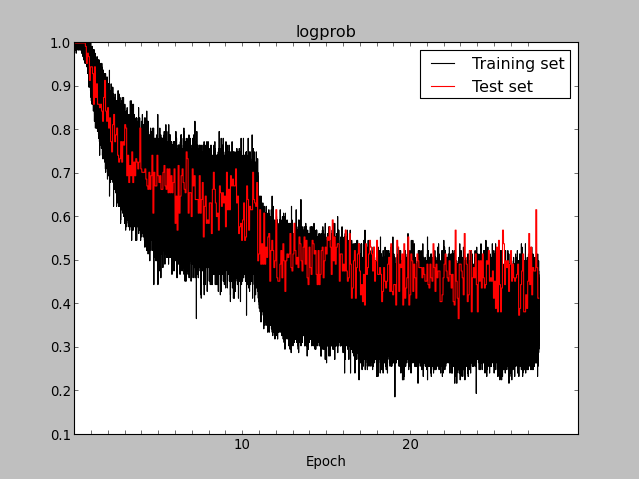}
    \caption{\small{Train set and test set error. Note that test set error reaches an average low around 45\% by 22 epochs.}}
    \label{fig:no_check_u600_n8}
\end{figure}

\subsection{Experiment 1}
Our first experiment is to test whether we can achieve similar performance to \cite{krizhevsky2012imagenet} with GPU A-SGD. We used the same settings we used in the single GPU cases, with $n_{sync} = 600$. For this experiment we use 8 GPU clients. The resulting learning curves are shown in (fig .\ref{fig:no_check_u600_n8}). We get near state of the art performance by epoch 22 which takes 3.3 days, before overfitting. This is about a 3.2x speed up. In our experience, the minibatch test set performance is usually 2-3\% higher than the overall test set performance after averaging 10 crops as in \cite{krizhevsky2012imagenet}. That is true here, the checkpoint before over-fitting gets a test error of 42.2\%.

\begin{figure}[h]
    \centering
    \includegraphics[width=0.65\textwidth]{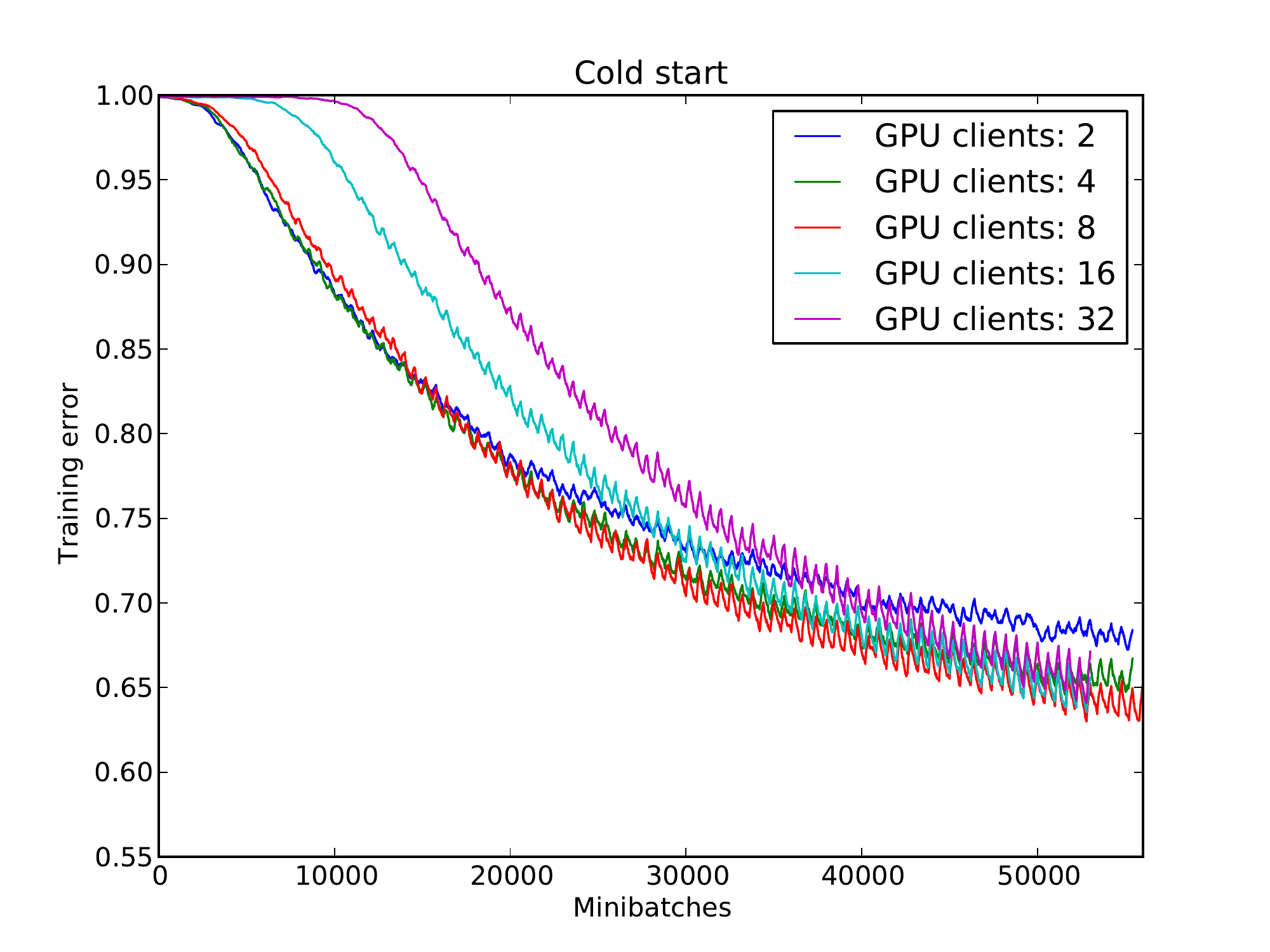}
    \caption{\small{Training error with a cold start. Notice early on training with 16 and 32 clients is much slower. Also notice that latter on, the 32 client GPU A-SGU model has the steepest learning curve.}}
    \label{fig:cold_start}
\end{figure}

For the next experiments we want to compare the speed up using varying numbers of GPU clients and varying values of $n_{sync}$. Since, it is hard to interpret many raw learning curves on a single plot, we smooth each plot using a sliding window of 400 mini batches. Also, we plot only the training error, so that the sliding window doesn't need to be adjusted for different values of $n_{sync}$. Since the training and testing error are very similar for the early training period we observe, we feel this is indicative of performance.

\subsection{Experiment 2}
In our second experiment, we examined the effect of a cold start on the learning, as the number of GPU clients increases from 2 to 32 (fig. \ref{fig:cold_start}). Each GPU A-SGD instance is run for 24 hours. We observe that as the number of GPUs increase, initial learning becomes much slower. We also observe that later in training, GPU A-SGD instances with more GPU clients learn more rapidly. We hypothesize that early in training, there are many gradient directions that may decrease error. Since each GPU client calculates different gradients, averaging them may slow progress. Later in training gradients become more consistent and averaging them increases the speed of learning. This result suggests that a warm start may be beneficial as suggested in \cite{NIPS2012_0598}. This may also be improved by methods that explicitly deal with variance in gradients such as adagrad \cite{duchi2011adaptive} and adadelta \cite{zeiler2012adadelta}.

\begin{figure}[h]
\centering
\subfloat{
  \includegraphics[width=65mm]{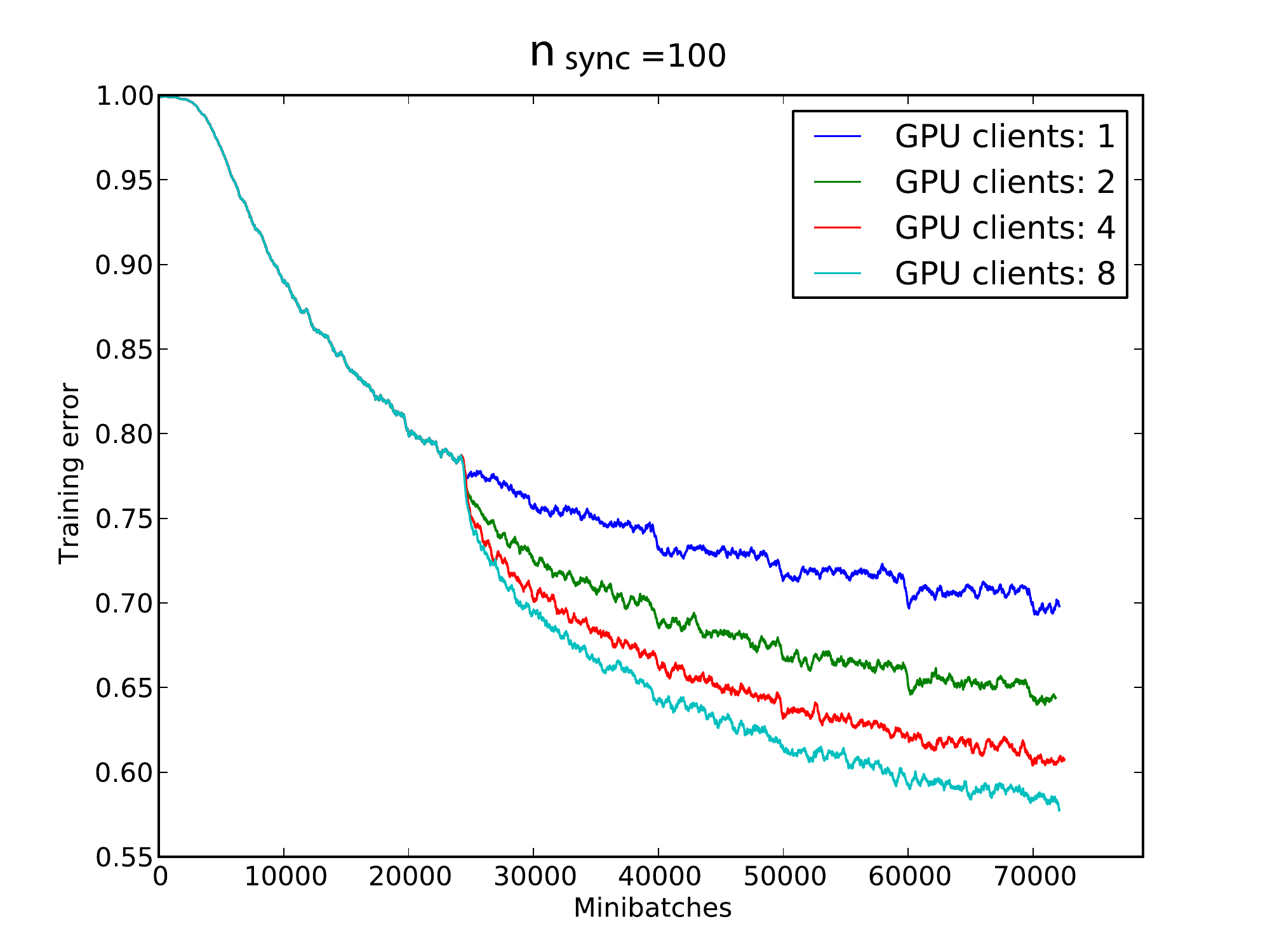}
}
\subfloat{
  \includegraphics[width=65mm]{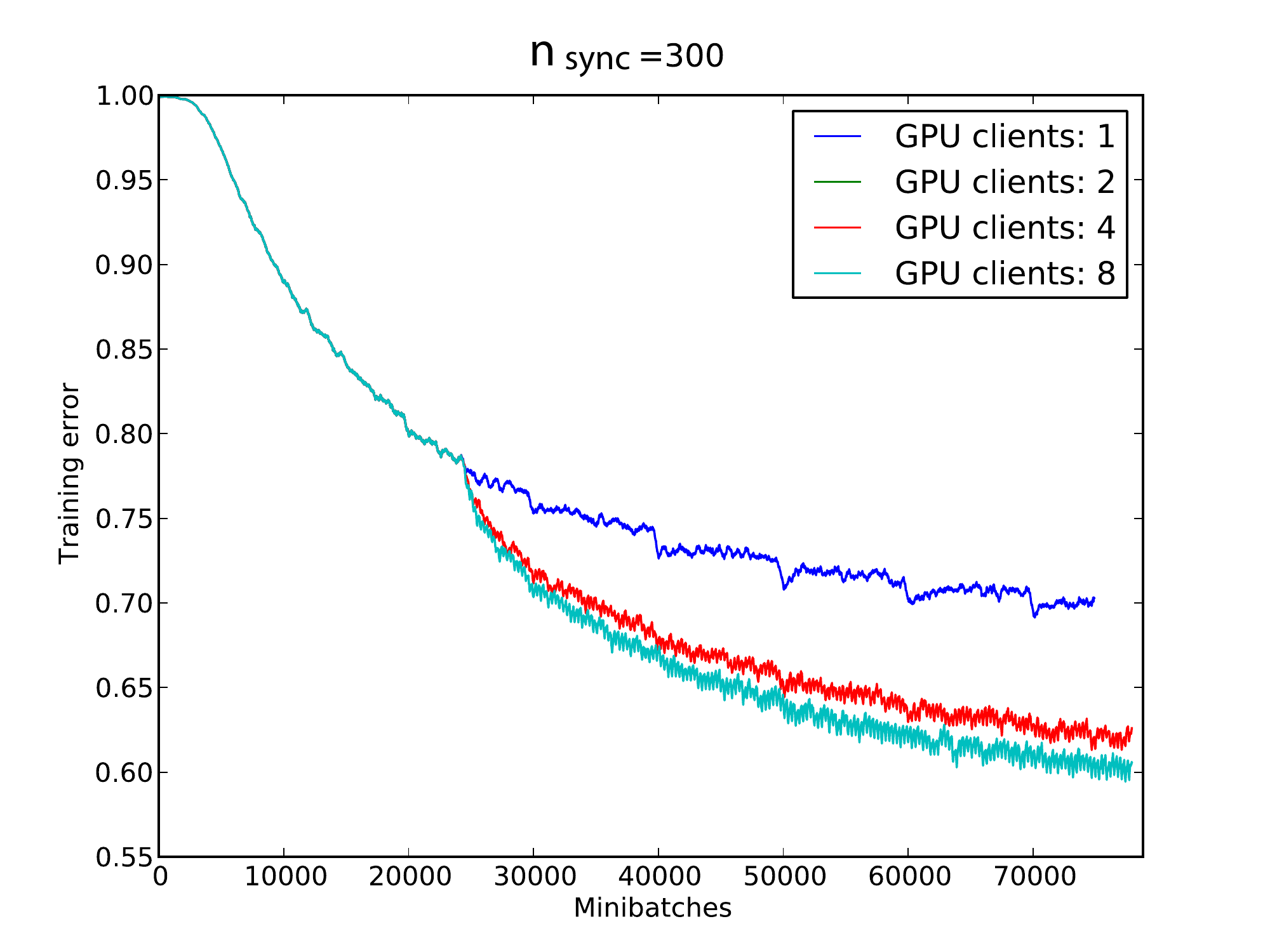}
}
\hspace{0mm}

\subfloat{
  \includegraphics[width=65mm]{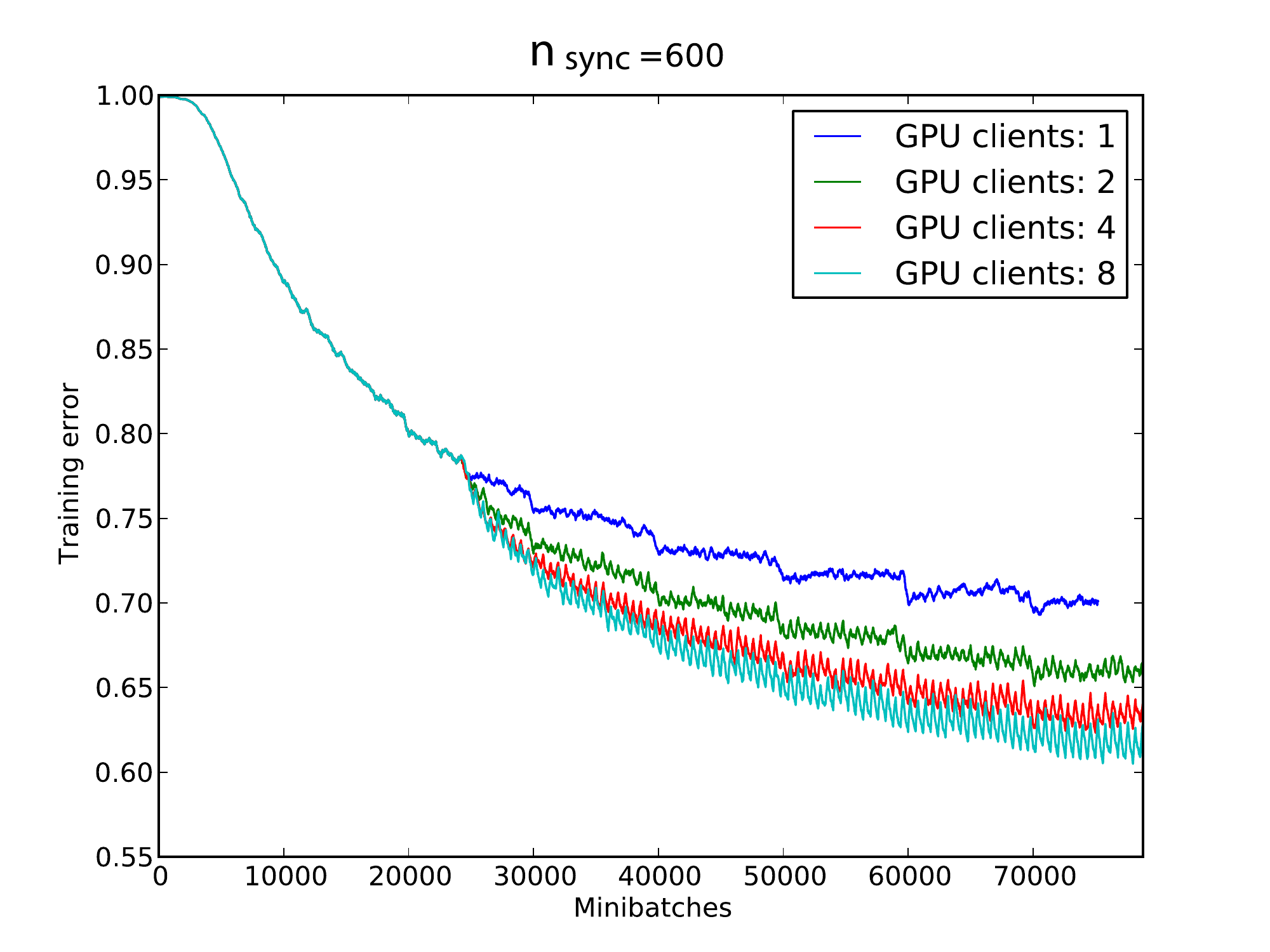}
}
\subfloat{
  \includegraphics[width=65mm]{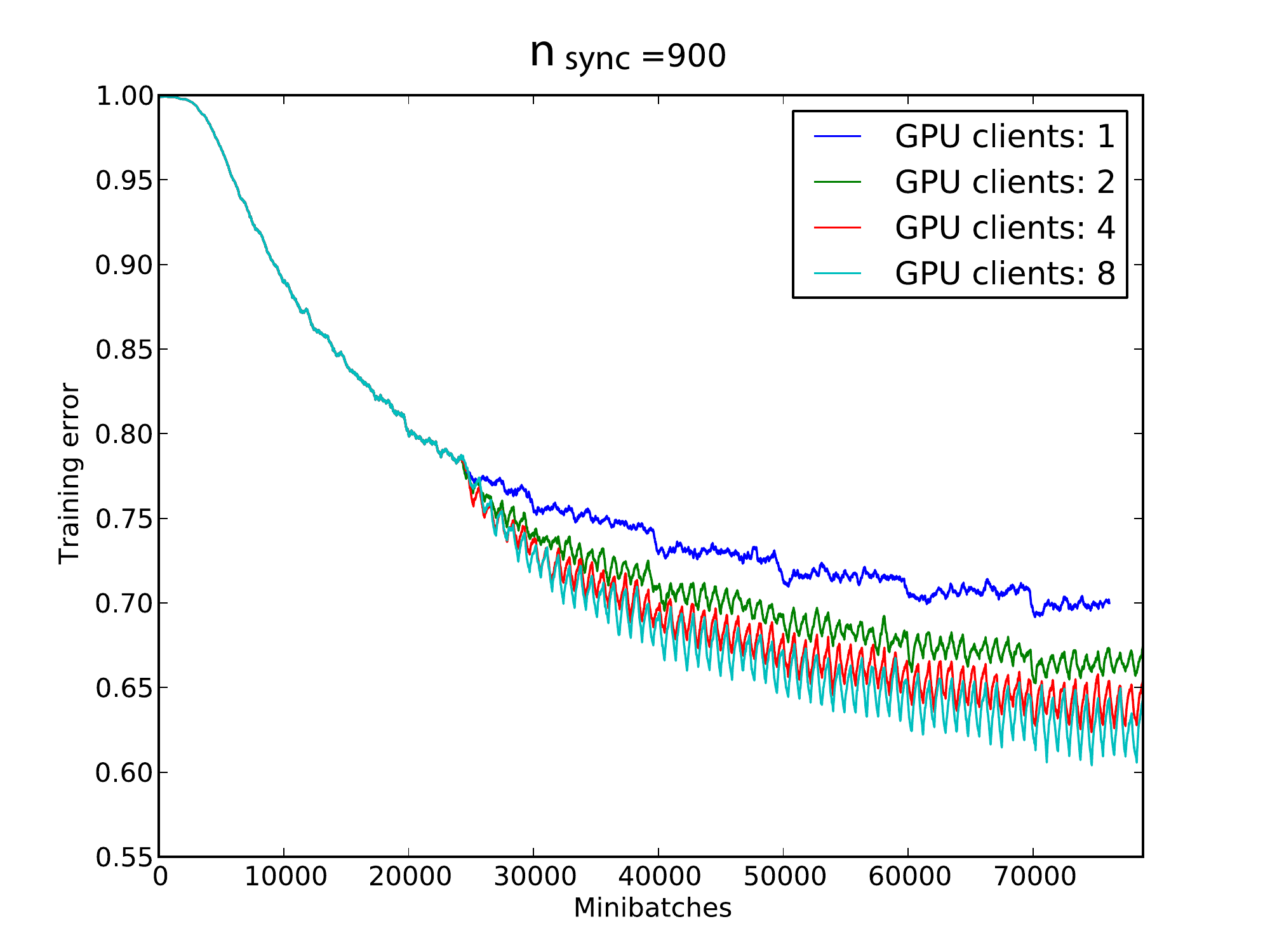}
}
\caption{\small{Training error with a warm start. Increasing the number of GPU client shows a significant speed up, across all values of $n_{sync}$. Note: for $n_{sync}=300$, the experiment for GPU cients=2 failed to run in time for this publication and it not included.}}
\label{fig:nsync}
\end{figure}

\begin{figure}[h]
    \centering
    \includegraphics[width=0.65\textwidth]{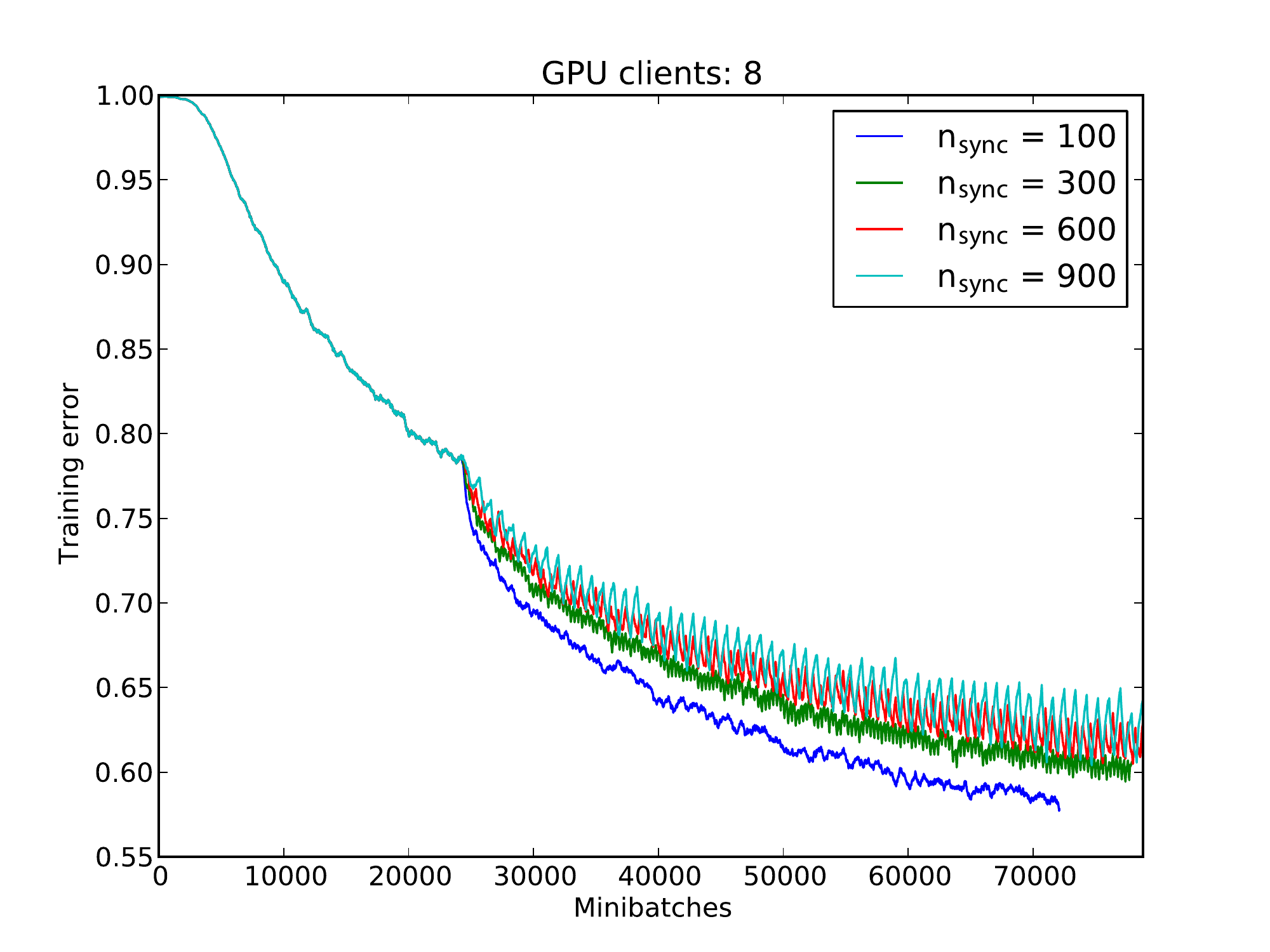}
    \caption{\small{Training error with a warm start, using 8 GPU clients. Notice that between $n_{sync}=900$ and $n_{sync}=100$ there is about a 4\% difference in training error.}}
    \label{fig:nsync_n8}
\end{figure}

\subsection{Experiment 3}
In our third experiment, we explore how $n_{sync}$ effects learning with many GPU clients. We try $n_{sync}$ values from 100 to 900 and 1-8 GPU clients (fig. \ref{fig:nsync}). We begin all experiments from a warm start, which we obtained by training the network on a single GPU for 12 hours. With a warm start, the effect of many GPU clients is clearer. When $n_{sync}=100$, our error decreases from 70\% with a single GPU to 58\% with 8 GPUs. Note that as $n_{sync}$ increases, the error curve has jagged artifacts. We believe these are from stale updates.

Also note that when $n_{sync}=100$, significantly fewer minibatches are processed in 24 hours, but the error rate is still lower. This suggests that while there is a cost associated with increased update frequency, it may still be a net win.

To emphasize these observations, we plot the learning curves for 8 GPU clients with $n_{sync}$ values from 100 to 900 (fig. \ref{fig:nsync_n8}).

\section{Future directions}

We plan to explore Adagrad \cite{duchi2011adaptive} and Adadelta \cite{zeiler2012adadelta} to see if they can further boost performance. We believe GPU A-SGD is a promising direction. Recently \cite{2013arXiv1311.2901Z} showed that larger models can further improve performance on computer vision tasks, and that these larger models begin to over fit, suggesting they would benefit from more training data. Both larger models, and larger training sets would benefit from faster training times.

%

\subsubsection*{Acknowledgments}

\small{
This material is based upon work supported by the National Science Foundation under Grant No. 392 NSF IIS 13-18971.

This research is part of the Blue Waters sustained-petascale computing project, which is supported by the National Science Foundation (award number OCI 07-25070) and the state of Illinois. Blue Waters is a joint effort of the University of Illinois at Urbana-Champaign and its National Center for Supercomputing Applications.
}

\small{
\bibliographystyle{ieee}
\bibliography{refs}
}
\end{document}